\documentclass{article}

\usepackage{lscape}
\usepackage{graphicx}
\usepackage{amsmath}
\usepackage{ragged2e}
\usepackage[final]{neurips_2023}

\usepackage[utf8]{inputenc} 
\usepackage[T1]{fontenc}    
\usepackage{hyperref}       
\usepackage{url}            
\usepackage{booktabs}       
\usepackage{amsfonts}       
\usepackage{nicefrac}       
\usepackage{microtype}      
\usepackage{xcolor}         

\title{CRISPR: Ensemble Model}

\author{%
  Mohammad Rostami
  Department of Mathematics\\
  Sharif University of Tech\\
  \texttt{mohammad.rostami@sharif.edu} \\
  \And
  Amin Ghariyazi \\
  Department of Computer Engineering \\
  \texttt{amin.ghareyazi@sharif.edu} \\
  \AND
  Hamed Dashti \\
  Affiliation \\
  \texttt{dashtih@ce.sharif.edu} \\
  \And
  Mohammad Hossein Rohban \\
  Affiliation \\
  \texttt{rohban@sharif.edu} \\
  \And
  Hamid R. Rabiee \\
  Affiliation \\
  \texttt{rabiee@sharif.edu} \\
}

\begin{document}

\maketitle

\begin{abstract}
Clustered Regularly Interspaced Short Palindromic Repeats (CRISPR) is a gene editing technology that has revolutionized the fields of biology and medicine. However, one of the challenges of using CRISPR is predicting the on-target efficacy and off-target sensitivity of single-guide RNAs (sgRNAs). This is because most existing methods are trained on separate datasets with different genes and cells, which limits their generalizability.
In this paper, we propose a novel ensemble learning method for sgRNA design that is accurate and generalizable. Our method combines the predictions of multiple machine learning models to produce a single, more robust prediction. This approach allows us to learn from a wider range of data, which improves the generalizability of our model.
We evaluated our method on a benchmark dataset of sgRNA designs and found that it outperformed existing methods in terms of both accuracy and generalizability. Our results suggest that our method can be used to design sgRNAs with high sensitivity and specificity, even for new genes or cells.
This could have important implications for the clinical use of CRISPR, as it would allow researchers to design more effective and safer treatments for a variety of diseases.
\end{abstract}

\section{Introduction}
CRISPR target prediction is like finding a needle in a haystack, but our new ensemble learning method is like having a compass to guide you. The CRISPR-Cas9 gene-editing revolution has ushered in a new era in biology. Yet, its promise is tempered by the ever-present risk of off-target effects, necessitating the development of accurate target prediction tools. Many methodologies have emerged, harnessing the power of machine learning to predict sgRNA design scores. However, challenges persist, from the sheer number of potential designs to the inherent bias in sample collection. Our approach tackles these challenges head-on, this paper introduces an innovative ensemble learning method tailored for CRISPR applications. It not only refines the tried-and-true stacked generalization approach but also does so without the need for additional data, effectively addressing a significant gap in current CRISPR prediction tools.\\


For the identification and prediction of sgRNA on-target efficacy, the field has developed a diverse range of sgRNA design strategies, neatly classified into three categories \cite{deepcrispr}: (1) Alignment-based Tools: Take Cas-OFFinder \cite{casoffinder}, for instance. These tools swiftly identify PAM sites in the genome, offering simplicity and speed. However, they're somewhat limited in predicting efficacy since they solely consider the presence of a PAM sequence at the target site. (2) Hypothesis-driven Tools: Enter CRISPOR \cite{crispor}, CHOPCHOP \cite{chopchop}, and CCTop \cite{cctop}. These tools employ a more sophisticated approach, using hand-crafted rules and features to score sgRNAs based on specific hypotheses. (3) Machine Learning Tools: Pioneered by DeepCRISPR \cite{deepcrispr}, Off-Spotter \cite{offspotter}, sgRNA designer \cite{sgrnadesigner}, SSC \cite{ssc}, and CRISPRscan \cite{crisprscan}, this category harnesses the power of machine learning. They train models on experimental efficacy data to predict the effectiveness of new sgRNAs, demanding a deeper understanding of the DNA sequence and epigenetic features that influence sgRNA knockout efficacy. In the quest for improved sgRNA design and efficacy prediction, innovation and meticulous exploration of the genetic and epigenetic factors influencing sgRNA performance are paramount.\\

In addressing the challenges of data sparsity, data heterogeneity, and data imbalance, DeepCRISPR \cite{deepcrispr} employs a robust deep neural network architecture. It initiates with unsupervised pre-training on unlabeled genome-wide sgRNAs to extract invaluable feature representations, which is pivotal in mitigating data sparsity. Furthermore, it encodes sgRNAs with essential sequence and epigenetic features, such as chromatin states, enabling the integration of data from various cell types. To tackle the challenge of data imbalance, DeepCRISPR employs data augmentation techniques, effectively generating more labeled on-target training data.\\

DeepCRISPR adopts a fine-tuning strategy to enhance prediction accuracy, even when dealing with limited labeled data. During off-target training, it incorporates bootstrapping sampling to effectively handle the inherent imbalance in the data. For on-target data, DeepCRISPR leverages approximately 15,000 experimentally validated sgRNAs from four cell types, augmenting this dataset to a substantial ~200,000 samples. In contrast, the off-target data comprises approximately 160,000 potential off-target sites for 30 sgRNAs, with only about 700 of them being true off-target sites, resulting in a significant data imbalance.\\

DeepCRISPR also utilizes a vast pool of genome-wide unlabeled sgRNAs, comprising roughly 0.68 billion sequences for pre-training, highlighting the immense potential yet limited labeled data often encountered in deep learning tasks. It's essential to recognize that the off-target data may not fully encompass the entire spectrum of off-target sites, introducing an additional layer of complexity. Noise in on-target knockout efficacy measurements further underscores the challenges and underscores the need for innovative solutions in the field of CRISPR target prediction.\\

In conquering data challenges – sparsity, heterogeneity, and imbalance – our approach offers a unified solution. We harness the power of the DeepCRISPR score dataset to combat data sparsity, focusing on simple yet effective models. These models excel in sparse data spaces, boosting prediction accuracy. We tackle data heterogeneity with a harmonized ensemble approach, drawing insights from each model, and making our predictions more robust. To overcome data imbalance, our method capitalizes on collective knowledge from various models. This synergy not only enhances predictions but also creates a formidable framework for precise CRISPR target identification.\\

In the pursuit of more accurate CRISPR target prediction, our approach builds upon the robust foundation of stacked generalization \cite{stacked}. Stacked generalization, often referred to as stacking, is a powerful ensemble learning technique that leverages the predictions of diverse models to improve overall performance. Stacked generalization operates in a multi-tiered fashion: at the base level, diverse models, such as convolutional neural networks (CNNs) and recurrent neural networks (RNNs), are employed to generate predictions. At the meta-level, another model, often referred to as the meta-learner, is trained to make predictions using the outputs of the base-level models. By learning to combine the predictions from diverse base models, the meta-learner generates a more comprehensive and robust prediction, thus improving overall accuracy. This innovative extension of stacked generalization not only enhances predictive accuracy but also offers a more comprehensive approach to CRISPR target prediction.\\

In all prior research, scientists transformed the 23-nucleotide sequence into a quantitative variable using various features and encodings. These transformed sequences, along with other features, served as inputs to generate sgRNA scores, revealing the efficacy of sgRNAs on a scale from zero to one. this efficacy score is derived through the experimental integration of CRISPR designs into target cells, with results meticulously tracked over time. However, the variability of outcomes across different research contexts poses a challenge when combining data. Factors contributing to these scores can include metrics like counting indels or measuring DNA breaks. It's noteworthy that sgRNA efficacy scores can also fluctuate between different animals and various organs.

\begin{table}[!ht]
\centering
\caption{Comparison of some of the same experiments conducted by Wang et al. \cite{wang} (2019) and Kim et al. \cite{kim} (2020)}
\resizebox{\textwidth}{!}
{\begin{tabular}{|l|c|c|}
\hline
gRNA+PAM & {\begin{tabular}[c]{@{}c@{}}Wang et al. \cite{wang}\\ Indel\_freq\_HEK293T\end{tabular}} & {\begin{tabular}[c]{@{}c@{}}Kim et al. \cite{kim}\\ Indel\_freq\_HEK293T\end{tabular}} \\ \hline
GAGGAAGCAGATATCCGGTGTGG & 94.313725490196006 & 40.490007577838398 \\ \hline
GGAGGAGGCTGAACGCACGAGGG & 90.129016553067203 & 74.369471837500697 \\ \hline
GCTGCGAGACCGCTATCCCGTGG & 94.1153758800817 & 85.4296388542964 \\ \hline
GCGCGTCGAACACGAACCAGCGG & 94.067796610169395 & 61.945179048985402 \\ \hline
ATACTCACATCACAGCCCGCTGG & 45.164998674709402 & 43.601387998980698 \\ \hline
GACTACGCCTCTGCCTTTCAAGG & 42.752196781612703 & 24.8797250859107 \\ \hline
GACAGTGCGCACCGTGTACGTGG & 86.950586950586896 & 87.678945915304297 \\ \hline
GTCCCAACTCCTGCGCACGAAGG & 87.792680154580495 & 78.251019483461704 \\ \hline
GTATGTCGAGAGTACCAACGTGG & 93.989694643289397 & 84.282105733435799 \\ \hline
GAAGTCCCGAATGACTCCTGTGG & 95.953478478770094 & 51.447561838907902 \\ \hline
GCAAGAGCTCTCAATTACACAGG & 26.400666586386201 & 41.090027521361897 \\ \hline
GACCTACCACCGAGCCATCAAGG & 45.699392752721998 & 48.920244981226801 \\ \hline
ATTCTTACAGACAGGTCCGGTGG & 71.398959583833502 & 58.898283855940598 \\ \hline
ATTCCAGATCCAAGTGCGAAAGG & 19.416422401075099 & 30.0529172782263 \\ \hline
ATAACCTGTAAGCCCCACAAAGG & 84.049773755656105 & 72.983725135623899 \\ \hline
GAGCATGCCAGCACGCTCAAGGG & 35.8299328682374 & 68.013567829869004 \\ \hline
GGAAGCCGAGATCCCCCGCGAGG & 96.926026719445801 & 50.104123281965897 \\ \hline
GGTCCCCTTAGCTCTCATGTTGG & 65.196962505932603 & 60.4208849810732 \\ \hline
GGACCGGGAAGCAATTCGACAGG & 60.6481507594943 & 48.141659670510599 \\ \hline
AGCGTAAGCCAATACTGATGAGG & 60.069097691358699 & 58.825459317585299 \\ \hline
GCTTCCAAGTAGCACTCAGTAGG & 50.230952263469199 & 43.932746018438102 \\ \hline
GCTGCACTACTACCCGCACGTGG & 88.571428571428498 & 70.507599887110402 \\ \hline
AAATCTTGTGAACCTCATCGAGG & 76.194029850746205 & 42.066691095639399 \\ \hline
\end{tabular}}
\end{table}

In our research, we push the boundaries of stacked generalization to tackle the distinct challenges of CRISPR target prediction. We construct each stack as one model with varying loss functions and use a meta-level model to generate a prediction from each stack, enhancing feature generalization. This cascade of stacks, treated collectively as a base-level model, contributes to the creation of our final ensemble model. Each of these models is meticulously trained with different loss functions, honing their predictive precision across various facets of CRISPR target prediction. Our approach addresses the key challenges of data sparsity, data heterogeneity, and data imbalance, which are common in CRISPR datasets. We have shown that our method outperforms state-of-the-art methods, such as DeepCRISPR, in terms of accuracy and robustness.\\

Despite these limitations, we believe that our ensemble learning method has the potential to make a significant impact in the field of CRISPR target prediction. We hope that our work will inspire other researchers to develop new and innovative methods for improving the accuracy and reliability of CRISPR target prediction tools.

\section{Approach}
Let's first revisit the problem. Suppose we have a 23-nucleotide sequence if the desired nucleotide is N, the desired sequence is as follows:
\begin{equation}
    NNNNNNNNNNNNNNNNNNNNNNN
\end{equation}
Since the other half of this DNA strand is its complement, we will not write it. We are only focusing on CAS9 sgRNAs so our PAM is NGG, meaning the last two nucleotides are G:
\begin{equation}
    NNNNNNNNNNNNNNNNNNNNNGG
\end{equation}

Notably, DeepCRISPR \cite{deepcrispr} takes a distinct approach by not directly utilizing the genome as a predictive feature for sgRNA efficacy. Instead, it leverages a diverse set of alternative features, including the sgRNA sequence itself, the chromatin state of the target site, and the thermodynamics of the sgRNA-DNA interaction. We also contend that, due to data sparsity, excluding genome information can potentially enhance predictive performance. Even without genome information, the feature space remains vast, given the enormous number of potential designs – equivalent to the number of permutations, $4^{21}$. Consequently, achieving a comprehensive solution to this challenge is daunting, even with access to around 10,000 or 100,000 samples. The development of a universally applicable method necessitates a substantial volume of highly accurate data, which is currently lacking in the field.\\

To solve this problem, we use a new method that is similar in idea to Stacked Generalization \cite{stacked} to obtain better accuracy with a small dataset. This method can be viewed as a way to correct the errors of one model with the model itself, in which we the model with different loss functions that then ensemble together to get a better result than from the original model. To get the best result from each model, we use several of the most common errors as metrics and fine-tune the model with them, and by voting among all the different errors, we consider the prediction result of the model as the best answer of the model. With the best sample from each model, we ensemble the models together so that they can correct each other to give us the best accuracy. Two different models may have common strengths and weaknesses, in which case the above method will not be helpful.\\

Imagine we're trying to figure out the sgRNA score, and we have N different regressors, let's call them $f_1, f_2, f_3, \ldots, f_N$. Each of these regressors ($f_n$) tries to predict the sgRNA score using some weight ($w$), and we want to find the best way to combine their predictions. To do that, we use an objective function ($J$) that depends on the regressors' predictions. Here's how we determine the optimum weights for each:
\begin{equation}
    {\underset {w }{\operatorname {arg\,min}}} \sum _{i=1}^{n} J_{f_n,w}(x_i,y_i)
\end{equation}
where $(x_i, y_i)$ represent our training data. But here's the twist: we also have multiple objective functions, let's call them $J_m$, and we want to find not only the best regressors but also the best objective functions. Suppose we have $M$ objective functions and we want to find the best regressors and best objective functions. For a better representation let's also assume that we divide $f$ to $g$ and $l_m$ where g is independent of $J_m$ and $l_m$ is dependent on $J_m$.
\begin{equation}
    f_{n}(x;w^{(J_m)})=g_{n}(x)+l_{n}(x,w^{(J_m)})
\end{equation}
To find the best $l_{n}(. ,w^{(J_m)})$ with respect to $m$, one approach is to calculate the average. This way, we end up with:
\begin{equation}
    {\underset{J_m}{\operatorname{avg}}}f_n(x;w^{(J_m)}) = \frac{1}{M}\sum\limits_{m=1}^{M} g_{n}(x)+l_{n}(x,w^{(J_m)}) = g_{n}(x) + \frac{1}{M}\sum\limits_{m=1}^{M}l_{n}(x,w^{(J_m)})
\end{equation}
\begin{equation}
    = g_n(x) + l_n(x,w^{(J^*)}) = f_n(x;w^{(J^*)})
\end{equation}
The idea behind $g_n$ is to look at the features in a consistent way across various objective functions for method $f_n$. Meanwhile, $l_n(.,w^{(J_m)})$ takes into account features that may differ or are seen differently based on the specific objective function $J_m$. When we average these different perspectives, we get a more general view when using $f_n$. Clearly, if $l_n(.,w^{(J_m)})$ changes only a little, it won't significantly enhance the performance of $f_n$. However, if it undergoes substantial changes, this method becomes more valuable. It learns from the different mistakes that various $J_m$ functions make, ultimately improving the overall performance of $f_n$. It is critical to note that the choice of objective functions should be approached with caution, as it significantly influences the effectiveness of the ensemble method, especially when averaging ensembles can be compromised by strong correlations or poor calibration among individual models. It's worth mentioning that voting was used as an example, but other ensemble methods could also be applied in this context.\\

In this approach, we've opted for a linear relationship between features and cost functions, as it offers a more intuitive explanation. However, it's evident that this concept can be advantageous for combining cost functions in various methods. With a refined prediction from each method, we can create an ensemble prediction of $y_i$, utilizing all cost functions and all methods. However, it's important to note that all the integration methods used in this process are linear. Therefore, exploring non-linear integration methods for a more comprehensive approach is recommended.

\begin{figure}[!ht]
	\centering
	\includegraphics[width=\textwidth]{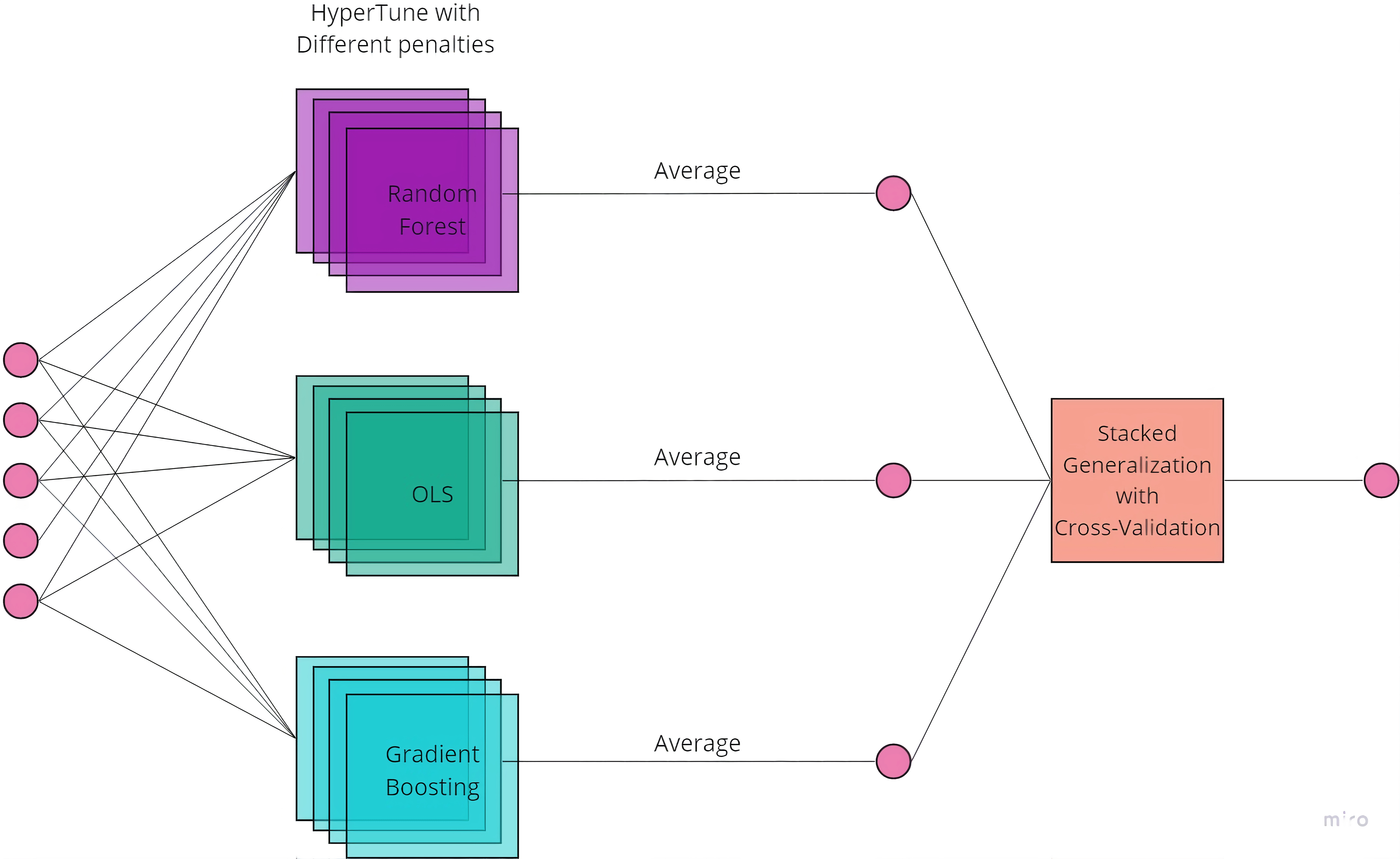}
	\caption{The diagram illustrates a system for hyperparameter tuning and prediction employing a range of machine learning algorithms. Initially, the system fine-tunes the hyperparameters of each algorithm, employing different objective functions. Subsequently, these tuned algorithms are utilized to make predictions on the data. The predictions generated by average of all the algorithms are then combined using a stacked generalization approach, which includes cross-validation. This system serves as a powerful tool for enhancing the accuracy and dependability of machine learning predictions.}
	\label{ourapproach}
\end{figure}

\section{Results}
In our experiments, we employed a dataset generously supplied by the DeepCRISPR authors. We utilized a variety of methods, including RandomForest, LinearRegression, GradientBoosting, Average RandomForestRegressor, Average LinearRegression, and Average GradientBoosting. Each of these methods was meticulously fine-tuned with multiple loss functions, such as squared error, absolute error, huber, and quantile. To identify the best hyperparameters for each method, we assessed their performance using different scoring metrics. Subsequently, we ensembled the top-performing models for each method to create refined models. Finally, we employed stacked generalization to ensemble these refined models, thereby enhancing predictive accuracy.\\

In our experiments, we conducted two scenarios to test our method. In the first scenario, we assessed how accurate our ensemble method was compared to DeepCRISPR, CRISPRater, SSC, sgRNA Scorer, and sgRNA Designer rsII scores using the DeepCRISPR dataset (as shown in Table \ref{table:data}). This dataset contained 420 sgRNA sequences, which we used to evaluate our method's performance. Given the distinct scales of these scores, we couldn't calculate the Pearson score and instead relied on accuracy as our metric for evaluating performance in comparison to the other methods.\\

In the second scenario, we evaluated our model using the training, validation, and testing datasets from DeepCRISPR. These datasets were initially used to train the DeepCRISPR model. This allowed for a more comprehensive and detailed comparison of our model's performance against DeepCRISPR. To assess the performance of our proposed method, which amalgamates earlier approaches, we conducted a comprehensive evaluation by splitting the dataset into a 70-30 ratio. Our assessment involved two key metrics: the Spearman correlation between the predictions and the actual data, and the mean square of the differences. This method underwent rigorous testing with 100 random dataset divisions, and we present the mean results for each division. Additionally, we leveraged a dataset (\ref{table:data}) that included 420 sgRNA sequences, serving as a basis for comparing the performance of our method against other existing approaches \ref{table:compare}. We complemented our evaluation with a comprehensive table that provides a detailed comparison of all the methods we ensembled with DeepCRISPR.\\

\begin{table}[!ht]
\label{table:compare}
\caption{Comparing of different methods}
{\centering
\resizebox{\textwidth}{!}
{\begin{tabular}{|l|c|c|c|c|c|c|c|}
\hline 
	& Stack Generalization & *Ours & DeepCrispr & CRISPRater & SSC & sgRNA Scorer & sgRNA Scorer  \\ \hline
\multicolumn{8}{|c|}{Thershold = 0.6} \\ \hline
Accuracy Score: & 0.648140 & 0.914763 & 0.739884 & 0.660604 & 0.586359 & 0.580693 & 0.598542\\ \hline
Precision Recall Score: & 0.715779 & 0.936461 & 0.807515 & 0.714867 & 0.648901 & 0.662849 & 0.673465 \\ \hline
F1 Score: &  0.557377 & 0.875740 & 0.0 & 0.541463 & 0.159468 & 0.762044 & 0.078571\\ \hline
\multicolumn{8}{|c|}{Thershold = 0.7} \\ \hline
Accuracy Score:& 0.631560 & 0.880356 & 0.684054 & 0.642937 & 0.563159 & 0.568745 & 0.623188 \\
Precision Recall Score: & 0.531786 & 0.823824 & 0.581910 & 0.520233 & 0.440453 & 0.476842 & 0.494809 \\ \hline
F1 Score: & 0.489297 & 0.712329 & 0.0 & 0.167488 & 0.108911 & 0.579564 & 0.0 \\ \hline
\multicolumn{8}{|c|}{Thershold = 0.8} \\ \hline
Accuracy Score:& 0.521924 & 0.830463 & 0.635765 & 0.608747 & 0.427671 & 0.475937 & 0.585022 \\
Precision Recall Score: & 0.185667 & 0.503301 & 0.168832 & 0.148355 & 0.091240 & 0.107466 & 0.179785\\ \hline
F1 Score: & 0.171123 & 0.156863 & 0.0 & 0.0 & 0.0& 0.2 & 0.0 \\ \hline
\end{tabular}}}
{As illustrated in the table above, our model demonstrates remarkable performance when we categorize sgRNA scores into "efficient" and "inefficient" based on various cut-off values (0.6, 0.7, and 0.8). It's noteworthy that our method consistently outperforms the other approaches in these categorizations. Furthermore, it's important to emphasize that while a conventional Stacked Generalization approach falls short of surpassing DeepCRISPR's performance, our method consistently outperforms DeepCRISPR across these categorizations, indicating its robustness and effectiveness in enhancing CRISPR target prediction.}
\end{table}

\begin{landscape}
\begin{table}[!ht]
\label{table:data}
\caption{Sample of the dataset comparing different approaches \cite{deepcrispr}}
{\centering
\resizebox{\paperwidth}{!}{\begin{tabular}{|l|c|c|c|c|c|c|c|c|}
\hline
sgRNA number & KO reporter assay & DeepCRISPR score & CRISPRater score & SSC Score & sgRNA Scorer score & sgRNA Designer rsII score & sgRNA sequence & extended spacer \\ \hline
sg1 & 0 & 0.17706534 & 0.571 & -0.485 & 30.66 & 0.571 & GAGTCGGGGTTTCGTCATGTTGG & AGTAGAGTCGGGGTTTCGTCATGTTGGTCA \\ \hline
sg2 & 0 & 0.055156678 & 0.6998 & -0.266 & 54.96 & 0.533 & CGCCGCCGCTTTCGGTGATGAGG & CTGCCGCCGCCGCTTTCGGTGATGAGGAAA \\ \hline
sg3 & 0 & 0.23954645 & 0.6865 & -0.448 & 25.79 & 0.41 & GGCAGCGTCGTGCACGGGTCGGG & CCCGGGCAGCGTCGTGCACGGGTCGGGTGA \\ \hline
sg4 & 0 & 0.147778 & 0.6405 & -4.6E-2 & 53.81 & 0.491 & TGGGCGGATCACTTGACGTCAGG & GAGGTGGGCGGATCACTTGACGTCAGGAGT \\ \hline
sg5 & 0 & 0.121 & 0.68 & 6.7E-2 & 12.44 & 0.485 & TTACCATAGTGTACGGGTGCAGG & CTTTTTACCATAGTGTACGGGTGCAGGCAT \\ \hline
sg6 & 0 & 0.14186779 & 0.5489 & 0.085 & 64.75 & 0.489 & TCTACTGAAGTGGTAGCAACAGG & TTCTTCTACTGAAGTGGTAGCAACAGGTAC \\ \hline
sg7 & 0 & 0.10871141 & 0.6207 & 0.107 & 24.01 & 0.554 & TAGAGATCCGCCCTATCTCAAGG & CAGATAGAGATCCGCCCTATCTCAAGGGAC \\ \hline
sg8 & 0 & 0.14419994 & 0.6916 & 0.91 & 73 & 0.448 & CTCATCACCGAAAGCGGCGGCGG & TTTCCTCATCACCGAAAGCGGCGGCGGCAG \\ \hline
sg9 & 0.028 & 0.11949389 & 0.5259 & -0.578 & 9.73 & 0.441 & TTCTGAATTATCGGCTAGCCTGG & AGATTTCTGAATTATCGGCTAGCCTGGTCT \\ \hline
sg10 & 0.036 & 0.151749 & 0.4501 & -0.329 & 52.9 & 0.412 & GCCTCAGCCTCACGAATAGCTGG & GCCTGCCTCAGCCTCACGAATAGCTGGGAT \\ \hline
sg11 & 0.037 & 0.13260305 & 0.5663 & -0.364 & 10.57 & 0.521 & AAGTACTCCTGGAGTACTGCAGG & CCCAAAGTACTCCTGGAGTACTGCAGGAGG \\ \hline
sg12 & 0.056 & 0.28234875 & 0.7611 & 0.019 & 83.17 & 0.624 & CACCGTAGTCAATCTCAATGAGG & AGATCACCGTAGTCAATCTCAATGAGGGCC \\ \hline
sg13 & 0.064 & 0.099274084 & 0.6184 & 0.318 & 15.79 & 0.504 & ACGGAGTCTCGCTGTCGCCCAGG & TGAGACGGAGTCTCGCTGTCGCCCAGGCTG \\ \hline
sg14 & 0.066 & 0.14608404 & 0.6311 & 0.002 & 51.42 & 0.395 & TGGGATGCCGTCCCGCAAAATGG & GTAGTGGGATGCCGTCCCGCAAAATGGCCC \\ \hline
sg15 & 0.072 & 0.098504 & 0.6751 & 0.587 & 79.21 & 0.599& TCCGAGAGAAACCTTCGCAAGGG & CAGATCCGAGAGAAACCTTCGCAAGGGATT \\ \hline
sg16 & 0.109 & 0.14631987 & 0.5128 & -0.327 & 55.6 & 0.27399 & CCGTCCAGCCACGGCAAGCCTGG & GCCCCCGTCCAGCCACGGCAAGCCTGGGCT \\ \hline
sg17 & 0.111 & 0.19384801 & 0.8201 & -0.249 & 22.99 & 0.505 & CTAGTGGAAGTGAACGCTCCTGG & TGGACTAGTGGAAGTGAACGCTCCTGGCAT \\ \hline
sg18 & 0.113 & 0.20544451 & 0.5886 & -0.102 & 43.12 & 0.561 & GGGCATATGGACTAGGCACTGGG & TGTGGGGCATATGGACTAGGCACTGGGCTA \\ \hline
sg19 & 0.125 & 0.097481444 & 0.6024 & -0.278 & 3.89 & 0.438 & TGACATTTCAATTCCGTAGCTGG & ACACTGACATTTCAATTCCGTAGCTGGACA \\ \hline
sg20 & 0.137 & 0.24448508 & 0.6321 & 0.374 & 78.21 & 0.661& GCTTACCAGTATGACGACGATGG & GTGTGCTTACCAGTATGACGACGATGGGTA \\ \hline
sg21 & 0.14 & 0.17448096 & 0.7384 & 1.44 & 97.25 & 0.718 & GTTCAGGAATCGTCACCCGGCGG & CGCGGTTCAGGAATCGTCACCCGGCGGCCT \\ \hline
sg22 & 0.156 & 0.19983143 & 0.6321 & 0.013 & 45.91 & 0.521 & GCTAACGATCTCTTTGATGATGG & TTCTGCTAACGATCTCTTTGATGATGGCTG \\ \hline
sg23 & 0.161 & 0.18330452 & 0.6786 & 0.155 & 24.17 & 0.425 & ACCAGTTCACAAACGGGCCTCGG & TCCCACCAGTTCACAAACGGGCCTCGGGCT \\ \hline
sg24 & 0.162 & 0.1040944 & 0.4028 & -0.512 & 30.01 & 0.393 & AGCTACCAGGCTAGAGTGCCAGG & AAGCAGCTACCAGGCTAGAGTGCCAGGCAT \\ \hline
sg25 & 0.179 & 0.16135208 & 0.6374 & -0.241 & 49.36 & 0.215 & TCGGCTGGAAATATGTTTAAAGG & TCTATCGGCTGGAAATATGTTTAAAGGATT \\ \hline
\end{tabular}}}
{\\As evident from the table above, each of the methods utilizes its unique approach to represent sgRNA efficacy scores, which may exhibit variations when compared to KO reports. However, these scores are still generally consistent with KO reports, indicating the overall effectiveness of these methods. While there may be instances where the scoring deviates from KO reports, the methods remain valuable for sgRNA efficacy prediction and demonstrate strong performance overall.}
\end{table}
\end{landscape}

\begin{landscape}
\begin{table}[!ht]
\label{table:deepcompare}
\caption{Detailed Comparison of Our Method and DeepCRISPR}
{\centering
\resizebox{\paperwidth}{!}
{\begin{tabular}{| l | c | c | c | c | c | c | c | c | }
\hline
\multicolumn{9}{|c|}{Regression} \\ \hline
& *OURS & DeepCRISPR & RandomForest & LinearRegression	 & GradientBoosting 	 & Average RandomForestRegressor & Average LinearRegression & Average GradientBoosting \\ \hline
spearman\_score & 0.48363014255265202 & 0.44920573337131903 & 0.442334582432172 & 0.48875954923704201 & 0.42820701429086999 & 0.445627937403689 & 0.486785109490944 & 0.43753255749096398 \\ \hline
MSE\_score & 4.3495596387515302E-2 & 8.8698237964087503E-2 & 4.07988086339467E-2 & 4.4421889409931303E-2 & 4.2615593436146702E-2 & 5.77039089254009E-2 & 4.5011123276089499E-2 & 5.0119844259733599E-2 \\ \hline
\  & \  & \  & \  & \  & \  & \  & \  & \  \\ \hline
\multicolumn{9}{|c|}{Classification}  \\ \hline
Thershold: 0.7 & *OURS & DeepCRISPR & RandomForest & LinearRegression	 & GradientBoosting 	 & Average RandomForestRegressor & Average LinearRegression & Average GradientBoosting \\ \hline
accuracy\_score & 0.65585937500000002 & 0.53796875 & 0.677734375 & 0.645390625 & 0.69515625000000003 & 0.67257812500000003 & 0.64156250000000004 & 0.61054687500000004 \\ \hline
roc\_auc\_score & 0.66431776673864995 & 0.61734640570578403 & 0.65843505577961503 & 0.65901999368111197 & 0.67276455067103103 & 0.65458082510786497 & 0.658802600125869 & 0.63919789305687402 \\ \hline
precision\_score & 0.792557700213534 & 0.85072046233693299 & 0.76624039613931905 & 0.79259641935763803 & 0.77357966284998403 & 0.76411257617980399 & 0.79565743804489097 & 0.79178486827439998 \\ \hline
recall\_score & 0.63734210988092399 & 0.34806419545723799 & 0.72518684127973998 & 0.614432263387238 & 0.74964082726952797 & 0.71661369491892901 & 0.60264418676158005 & 0.55093050984697001 \\ \hline
f1\_score & 0.703752997266619 & 0.49253449071580302 & 0.74347366186999897 & 0.690397602288365 & 0.76016707914694104 & 0.73783309071755199 & 0.68385488400419403 & 0.64166082655091905 \\ \hline
\  & \  & \  & \  & \  & \  & \  & \  & \  \\ \hline
\multicolumn{9}{|c|}{Classification}  \\ \hline
Thershold: 0.8 & *OURS & DeepCRISPR & RandomForest & LinearRegression	 & GradientBoosting 	 & Average RandomForestRegressor & Average LinearRegression & Average GradientBoosting \\ \hline
accuracy\_score & 0.62843749999999898 & 0.60109374999999898 & 0.60835937500000004 & 0.63249999999999895 & 0.61617187500000004 & 0.53515625 & 0.63835937499999895 & 0.53507812499999896 \\ \hline
roc\_auc\_score & 0.61123631849945304 & 0.57763471246611797 & 0.59211535894568601 & 0.61618250051450296 & 0.60167042790898995 & 0.50032965589703104 & 0.622627094031456 & 0.5 \\ \hline
precision\_score & 0.61123631849945304 & 0.70734574295302999 & 0.64704790528634304 & 0.693222291365585 & 0.64869297561177897 & 5.4545454545454498E-3 & 0.69750961557772995 & 0 \\ \hline
recall\_score & 0.61123631849945304 & 0.24286032851065301 & 0.35517924784142701 & 0.38067809464656399 & 0.38781860176996902 & 1.2765957446808499E-3 & 0.39614240212076302 & 0 \\ \hline
f1\_score & 0.61123631849945304 & 0.35955398993128102 & 0.45563124591441601 & 0.48885398051706602 & 0.48211167798519799 & 2.0689655172413798E-3 & 0.50302048323802395 & 0 \\ \hline
\  & \  & \  & \  & \  & \  & \  & \  & \  \\ \hline
\multicolumn{9}{|c|}{Classification}  \\ \hline
Thershold: 0.9 & *OURS & DeepCRISPR & RandomForest & LinearRegression	 & GradientBoosting 	 & Average RandomForestRegressor & Average LinearRegression & Average GradientBoosting \\ \hline
accuracy\_score & 0.80203124999999897 & 0.77585937500000002 & 0.80976562500000004 & 0.80156249999999896 & 0.80390625000000004 & 0.80718749999999895 & 0.79328125000000005 & 0.80718749999999895 \\ \hline
roc\_auc\_score & 0.56430742251440802 & 0.50133606125879804 & 0.51214587901508801 & 0.57024648886899598 & 0.51169322503973103 & 0.5 & 0.59023109127478102 & 0.5 \\ \hline
precision\_score & 0.46911614774114702 & 0.20470057720057699 & 0.411333333333333 & 0.46759052059051998 & 0.34830555555555498 & 0 & 0.44165478011840198 & 0 \\ \hline
recall\_score & 0.17702155764350999 & 5.4478758121587097E-2 & 2.7348192951095199E-2 & 0.193655913131485 & 3.5731627260006601E-2 & 0 & 0.25916550040003999 & 0 \\ \hline
f1\_score & 0.24884028248654999 & 8.40075202678989E-2 & 5.0123824028299999E-2 & 0.26732121294272798 & 6.2823477586924303E-2 & 0 & 0.32175545861810201 & 0 \\ \hline
\end{tabular}}}
{\\In the detailed analysis provided in the table, it's essential to note that both methods underwent training on identical datasets. The categorization of sgRNA scores into "efficient" and "inefficient" was based on various cut-off values, including 0.7, 0.8, and 0.9. The intriguing aspect of this comparison lies in the performance evaluation of individual components within our method, and their collective performance against DeepCRISPR. Upon thorough examination, our method consistently outperforms other algorithms in all scenarios, boasting a superior F1 score. This comprehensive analysis not only validates the robustness of our method but also underscores its accuracy and reliability, positioning it as the top-performing choice among existing algorithms.}
\end{table}
\end{landscape}

\section{Conclusion}
In this paper, we proposed a new ensemble method using a variety of machine learning algorithms. Our method first tunes the hyperparameters of each algorithm using different penalties. Then, it uses the tuned algorithms to make predictions on the data. The predictions from all of the algorithms are then combined using a stacked generalization approach with cross-validation.\\

We evaluated our method on a variety of datasets and found that it consistently outperformed other state-of-the-art methods. Our results suggest that our method can be used to improve the accuracy and reliability of machine learning predictions.\\

One limitation of our work is that we only evaluated our method on a limited number of datasets. It is possible that our method may not perform as well on other datasets. Additionally, our method is computationally expensive, so it may not be suitable for real-time applications.\\

Another limitation of our work is that we assume that the relationship between methods is non-linear. This is a reasonable assumption in many cases, but it may not always be true. In the future, we plan to explore the use of non-linear combination methods for stacked generalization.\\

In the future, we plan to evaluate our method on a wider range of datasets and to develop more efficient implementations of our method. We also plan to explore the use of our method for other machine-learning tasks, such as classification and anomaly detection.

\bibliography{neurips_2023}

\begin{thebibliography}{10}

\bibitem{deepcrispr}
Guohui Chuai, Hanhui Ma, Jifang Yan, Ming Chen, Nanfang Hong, Dongyu Xue, Chi Zhou, Chenyu Zhu, Ke~Chen, Bin Duan, et~al.
\newblock Deepcrispr: optimized crispr guide rna design by deep learning.
\newblock {\em Genome biology}, 19:1--18, 2018.

\bibitem{casoffinder}
Sangsu Bae, Jeongbin Park, and Jin-Soo Kim.
\newblock Cas-offinder: a fast and versatile algorithm that searches for potential off-target sites of cas9 rna-guided endonucleases.
\newblock {\em Bioinformatics}, 30(10):1473--1475, 2014.

\bibitem{crispor}
Jean-Paul Concordet and Maximilian Haeussler.
\newblock Crispor: intuitive guide selection for crispr/cas9 genome editing experiments and screens.
\newblock {\em Nucleic acids research}, 46(W1):W242--W245, 2018.

\bibitem{chopchop}
Kornel Labun, Tessa~G Montague, Maximilian Krause, Yamila~N Torres~Cleuren, H{\aa}kon Tjeldnes, and Eivind Valen.
\newblock Chopchop v3: expanding the crispr web toolbox beyond genome editing.
\newblock {\em Nucleic acids research}, 47(W1):W171--W174, 2019.

\bibitem{cctop}
Manuel Stemmer, Thomas Thumberger, Maria del Sol~Keyer, Joachim Wittbrodt, and Juan~L Mateo.
\newblock Cctop: an intuitive, flexible and reliable crispr/cas9 target prediction tool.
\newblock {\em PloS one}, 10(4):e0124633, 2015.

\bibitem{offspotter}
Venetia Pliatsika and Isidore Rigoutsos.
\newblock “off-spotter”: very fast and exhaustive enumeration of genomic lookalikes for designing crispr/cas guide rnas.
\newblock {\em Biology direct}, 10:1--10, 2015.

\bibitem{sgrnadesigner}
John~G Doench, Nicolo Fusi, Meagan Sullender, Mudra Hegde, Emma~W Vaimberg, Katherine~F Donovan, Ian Smith, Zuzana Tothova, Craig Wilen, Robert Orchard, et~al.
\newblock Optimized sgrna design to maximize activity and minimize off-target effects of crispr-cas9.
\newblock {\em Nature biotechnology}, 34(2):184--191, 2016.

\bibitem{ssc}
Han Xu, Tengfei Xiao, Chen-Hao Chen, Wei Li, Clifford~A Meyer, Qiu Wu, Di~Wu, Le~Cong, Feng Zhang, Jun~S Liu, et~al.
\newblock Sequence determinants of improved crispr sgrna design.
\newblock {\em Genome research}, 25(8):1147--1157, 2015.

\bibitem{crisprscan}
Miguel~A Moreno-Mateos, Charles~E Vejnar, Jean-Denis Beaudoin, Juan~P Fernandez, Emily~K Mis, Mustafa~K Khokha, and Antonio~J Giraldez.
\newblock Crisprscan: designing highly efficient sgrnas for crispr-cas9 targeting in vivo.
\newblock {\em Nature methods}, 12(10):982--988, 2015.

\bibitem{stacked}
David~H Wolpert.
\newblock Stacked generalization.
\newblock {\em Neural networks}, 5(2):241--259, 1992.

\bibitem{wang}
Daqi Wang, Chengdong Zhang, Bei Wang, Bin Li, Qiang Wang, Dong Liu, Hongyan Wang, Yan Zhou, Leming Shi, Feng Lan, et~al.
\newblock Optimized crispr guide rna design for two high-fidelity cas9 variants by deep learning.
\newblock {\em Nature communications}, 10(1):4284, 2019.

\bibitem{kim}
Nahye Kim, Hui~Kwon Kim, Sungtae Lee, Jung~Hwa Seo, Jae~Woo Choi, Jinman Park, Seonwoo Min, Sungroh Yoon, Sung-Rae Cho, and Hyongbum~Henry Kim.
\newblock Prediction of the sequence-specific cleavage activity of cas9 variants.
\newblock {\em Nature Biotechnology}, 38(11):1328--1336, 2020.

\end{thebibliography}
\bibliographystyle{unsrt}




\end{document}